\documentclass{article}

\usepackage{microtype}
\usepackage{graphicx}
\usepackage{subfigure}
\usepackage{graphicx}
\usepackage{amsmath}
\usepackage{amssymb}
\usepackage{booktabs}
\usepackage{makecell}
\usepackage{multicol}
\usepackage{multirow}
\usepackage{color}
\usepackage{xcolor}
\usepackage{colortbl,booktabs} 

\usepackage[ruled, lined,  commentsnumbered, longend]{algorithm2e}

\SetCommentSty{mycommfont}
\usepackage[pagebackref,breaklinks,colorlinks]{hyperref}

\usepackage[capitalize]{cleveref}
\crefname{section}{Sec.}{Secs.}
\Crefname{section}{Section}{Sections}
\Crefname{table}{Table}{Tables}
\crefname{table}{Tab.}{Tabs.}

\usepackage[accepted]{icml2023}

\usepackage{amsmath}
\usepackage{amssymb}
\usepackage{mathtools}
\usepackage{amsthm}

\theoremstyle{plain}

\theoremstyle{definition}

\theoremstyle{remark}

\usepackage[textsize=tiny]{todonotes}

\icmltitlerunning{Instant Soup: Cheap Pruning Ensembles in A Single Pass Can Draw Lottery Tickets from Large Models}

\begin{document}

\twocolumn[
\icmltitle{Instant Soup: Cheap Pruning Ensembles in A Single Pass Can \\ Draw Lottery Tickets from Large Models}



\icmlsetsymbol{equal}{*}

\begin{icmlauthorlist}
\icmlauthor{Ajay Jaiswal}{yyy}
\icmlauthor{Shiwei Liu}{yyy}
\icmlauthor{Tianlong Chen}{yyy}
\icmlauthor{Ying Ding}{yyy}
\icmlauthor{Zhangyang Wang}{yyy}

\end{icmlauthorlist}

\icmlaffiliation{yyy}{University of Texas at Austin}

\icmlcorrespondingauthor{Ajay Jaiswal}{ajayjaiswal@utexas.edu}

\icmlkeywords{Machine Learning, ICML}

\vskip 0.3in
]



\printAffiliationsAndNotice{\icmlEqualContribution} 

\begin{figure}[h]
  \centering
  \includegraphics[width=\linewidth]{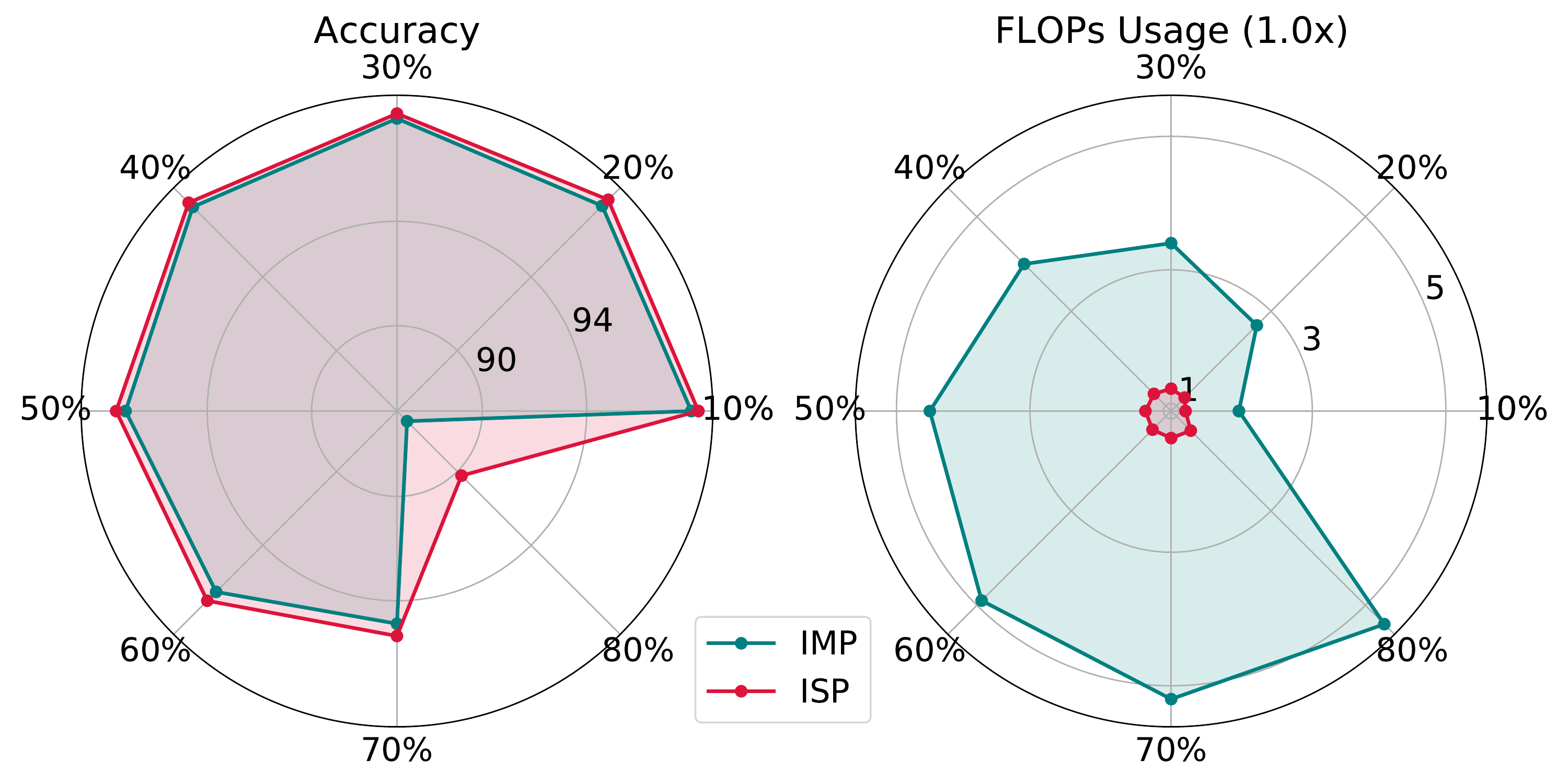}
  \vspace{-0.4cm}
  \caption{Fine-tuning Accuracy ($\uparrow$) and FLOPs counts ($\downarrow$) of subnetworks with sparsity $s\in\{10\%, 20\%, ..., 80\%\}$ on $\texttt{CIFAR-10}$ test set with a pre-trained CLIP (\texttt{ViT-B32}) model checkpoint. Note that ISP requires approximately single-pass computational cost of IMP to generate subnetworks with better performance than multiple rounds of IMP as required in Lottery Tickets.}
  \label{fig:teaure}
\end{figure}

\begin{abstract}
Large pre-trained transformers have been receiving explosive attention in the past few years, due to their wide adaptability for numerous downstream applications via fine-tuning, but their exponentially increasing parameter counts are becoming a primary hurdle to even just fine-tune them without industry-standard hardware. Recently, Lottery Ticket Hypothesis (LTH) and its variants, have been exploited to prune these large pre-trained models generating subnetworks that can achieve similar performance as their dense counterparts, but LTH pragmatism is enormously inhibited by repetitive full training and pruning routine of iterative magnitude pruning (IMP) which worsens with increasing model size. Motivated by the recent observations of model soups, which suggest that fine-tuned weights of multiple models can be merged to a better minima, we propose \textbf{Instant Soup Pruning (ISP)} to generate lottery ticket
quality subnetworks, using a fraction of the original IMP cost by replacing the expensive intermediate pruning stages of IMP with computationally efficient weak mask generation and aggregation routine. More specifically, during the mask generation stage, ISP takes a small handful of iterations using varying training protocols and data subsets to generate many weak and noisy subnetworks, and superpose them to average out the noise creating a high-quality denoised subnetwork. Our extensive experiments and ablation on two popular large-scale pre-trained models: \texttt{CLIP} (unexplored in pruning till date) and \texttt{BERT} across multiple benchmark vision \texttt{\{MNIST, SVHN, Cars, GTSRB, CIFAR-10, CIFAR-100\}} and language datasets \texttt{\{MNLI, QNLI, QQP, SST, ...\}} validate the effectiveness of ISP compared to several state-of-the-art pruning methods. Additionally, we show that ISP can be easily modified with minimal overhead to produce benefits comparable to model soups, without the prerequisite to generate multiple candidates fine-tuned models. Codes are available at:  \url{https://github.com/VITA-Group/instant\_soup}.

\end{abstract}

\vspace{-0.8cm}
\section{Introduction}
\label{sec:intro}


Large-scale transfer learning has recently become \textit{show-stealer} in modern deep learning, and transformer-based pre-trained models \cite{devlin2018bert,liu2019roberta,dosovitskiy2020image,liu2021swin,radford2021learning} are now achieving state-of-the-art performance for a wide array of real-world computer vision \cite{dosovitskiy2020image,Han2020ASO,li2023cancergpt, Touvron2021TrainingDI,Mao2022SingleFA,jaiswal2021scalp,Zheng2021EndtoEndOD,Parmar2018ImageT} and natural language processing \cite{yang2019xlnet,liu2019roberta,talmor2018commonsenseqa,Jaiswal2021RadBERTCLFC,zheng2023outline,yang2019end,wang2018glue,ding2019cognitive,chowdhery2022palm,wei2022chain,jaiswal2023attend} applications. With the astonishing explosion of parameter counts (millions to billions) in the past few years, while chasing performance gains, fine-tuning these large pre-trained models with non-industry standard hardware is becoming seemingly impossible, in addition to expensive inference and steep environmental cost. In the hustle of building gigantic models, a parallel and growing field of model compression has been exploring the prospects to compress these enormous models at the cost of marginal/no sacrifice in performance, effectively reducing their computational and memory footprints.

\begin{figure*}
  \centering
  \includegraphics[width=0.9\linewidth]{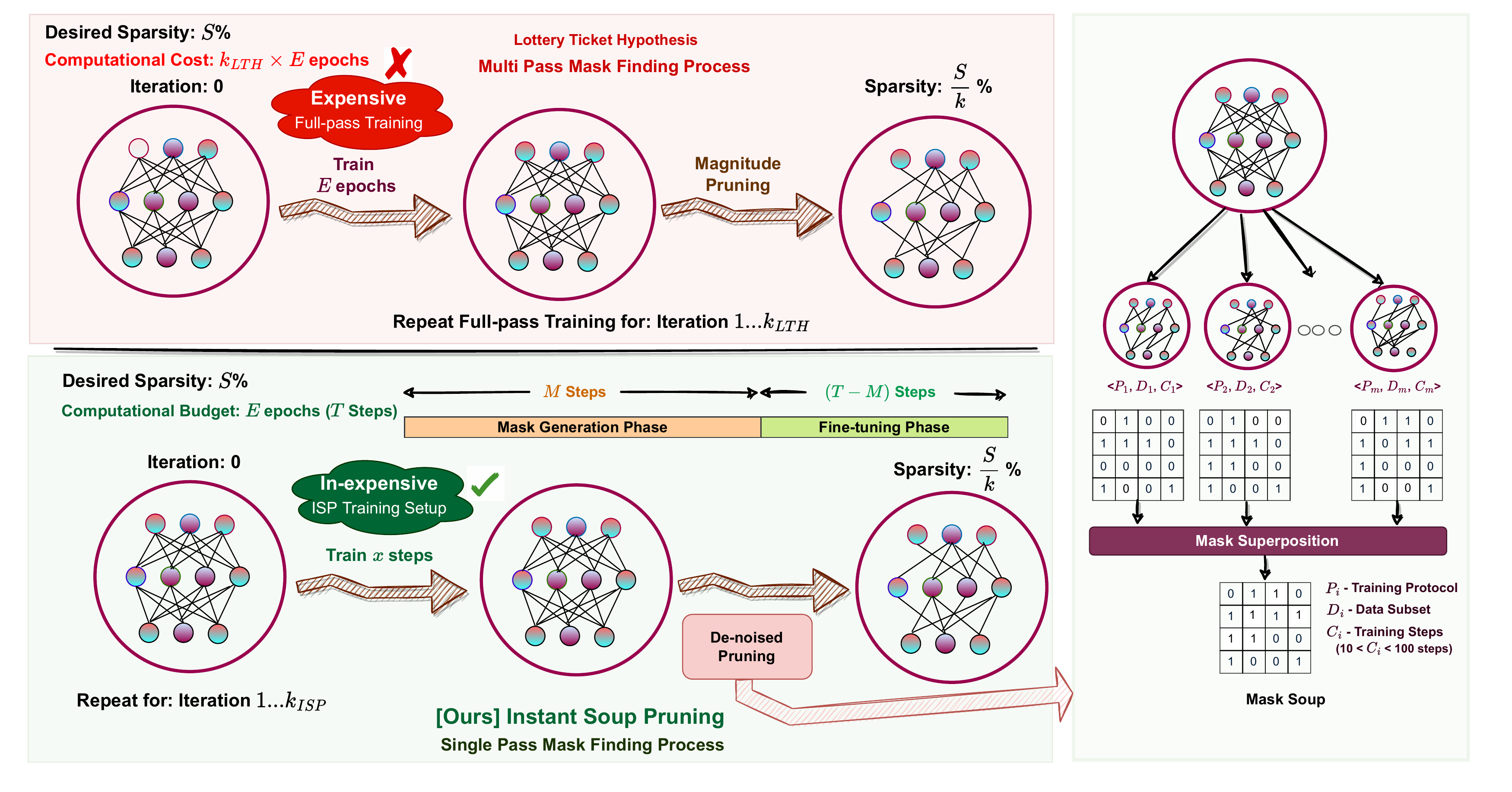}
  \vspace{-1.5em}
  \caption{Overview of our proposed Instant Soup Pruning. We provide a detailed illustration of our proposed technique compared to conventional LTH with IMP. ISP replaces the expensive intermediate pruning stages of IMP with computationally-cheap weak mask generation and denoising while outperforming LTH. Unlike LTH, ISP consumes computation budget equivalent of a single pass of LTH.}
  \label{fig:main_architecture}
  \vspace{-1em}
\end{figure*}

Among many efforts for compressing models and accelerating inference \cite{frankle2018lottery,chen2020lottery,jaiswal2022spending,yin2022lottery,jaiswal2022training,lee2018snip,yu2017compressing,yu2020gradient,fang2023depgraph,chen2023sparse,jaiswal2022ros,liu2023sparsity}, network pruning eliminates unnecessary weights to generate smaller \textit{subnetworks} in place of dense networks attaining similar performance, stands out as one of the most effective techniques. Lottery Ticket Hypothesis (LTH) \cite{frankle2018lottery} and its variants, reveal that dense, randomly-initialized networks contain small subnetworks which can match the test accuracy of original networks. Despite their insightful findings, there still exists a large \textbf{gap} in the practicality of these methods because of the fully dense training routine of IMP, which exacerbates with an increase in model capacity. Recently proposed EarlyBird routine \cite{you2019drawing,chen2020earlybert} which attempts to draw the winning tickets early in training, and pruning at initialization techniques have shown some promise in mitigating search through an expensive and tedious iterative process, yet their effectiveness for the large-scale pre-trained network is highly under-explored or they tend to have sub-standard performance at non-trivial sparsities. In this work, we ask: \textit{Does there exist a principled and cheaper approach for fastly drawing high-quality lottery tickets in large pre-trained models within a limited computational budget, while preserving its performance and transferability?}

To this end, we explore the feasibility of obtaining cheap tickets for popular large pre-trained models CLIP\cite{radford2021learning} and BERT\cite{devlin2018bert} constrained by no multi-pass repetitive full-training, to meet the permissible computational budget. One straightforward approach is to curtail per round training cost of iterative magnitude pruning (IMP) with early stopping, but we found that such non-careful pruning approach often produces highly variable and substandard tickets, presumably due noisy and unstable state of the network, when pruned. Our close analysis of the pruned mask generated by LTH and cheap pruning methods (one-shot pruning) found that large-scale pre-trained transformers are highly overparameterized and pruning them at trivial sparsities does not necessarily require an expensive LTH paradigm, which encourages us to start pruning nonchalantly at trivial sparsities and become scrupulous at non-trivial sparsities. In addition, recently several works \cite{wortsman2022model,ilharco2022patching,juneja2022linear} investigate the intriguing phenomenon of ``model soups", and have shown that weights of multiple dense fine-tuned models can be merged together into better solutions lying in low error basins. In the sparse setting, a very recent attempt \cite{yin2022lottery} reused the byproduct of IMP, and showed that tickets generated at each iteration of IMP could be superposed into a stronger subnetwork. However, its observations are limited to small-scale networks, and more importantly, the algorithm does not contribute to the computational efficiency of either finding lottery tickets or (re-)training networks. Motivated by these observations, we are interested to investigate: \textit{if we can leverage the soup observations to eliminate noise induced by early pruning in IMP iteration, effectively leading to the stable sparse subnetwork and reduced cost}. 

We propose \textbf{Instant Soup Pruning (ISP)}, a model soup-inspired perspective dedicated to generating lottery ticket quality subnetworks, using a fraction of the original IMP cost. More specifically, ISP uses a miniature random subset of training data to generate many weak and noisy subnetworks, and superpose them to average out the noise creating a \underline{high-quality denoised subnetwork}. Similar to traditional IMP, ISP repeats the denoising routine following well-managed training iterations till the desired sparsity is reached, eliminating the need of IMP to perform a full pass of training before every pruning routine. Our experiments on CLIP (unexplored in pruning literature till date) and BERT across multiple datasets illustrate that ISP can find sparse subnetworks with better quality than LTH, with an affordable cost no more than a single pass of IMP. In addition to dense-to-sparse paradigm, interestingly, our ISP routine can be bluntly incorporated in dense-to-dense paradigm to incorporate the benefits of model soups in dense pre-trained models at almost negligible training cost at the pre-training stage, ultimately leading to better-fine tuning performance. Our contributions can be summarized as:

\vspace{-0.3cm}
\begin{itemize}
    \item [$\blacksquare$] We propose \textbf{Instant Soup Pruning (ISP)}, a novel pruning strategy that seamlessly integrates the ``model soup" idea to significantly reduce the computational cost of IMP. ISP replaces the expensive intermediate pruning stages of IMP with computationally-cheap weak mask generation and denoising, while outperforming IMP for large pre-trained models.
    \vspace{-0.1cm}
    \item [$\blacksquare$] ISP naturally provides a ``self-denoising" ability to eliminate the necessity of generating high quality/expensive masks (presumably ``good solution basin”) at each pruning stage, by instead generating multiple computationally inexpensive weak masks and averaging them out to reduce their solution noise. 
    \vspace{-0.1cm}
    \item [$\blacksquare$] In the dense-to-dense paradigm, ISP can be adapted to \textbf{Instant Model Soup}, to inject the benefits of model soups in dense pre-trained models at marginal training cost, thereby improving fine-tuning performance comparable to model soups without the need to generate multiple fine-tuned models.
    \vspace{-0.1cm}
    \item [$\blacksquare$] Our extensive experiments on two popular large-scale pre-trained models (CLIP \& BERT) across multiple benchmark vision \texttt{\{MNIST, SVHN, Cars, GTSRB, CIFAR-10, CIFAR-100\}} \& language datasets \texttt{\{MNLI, QQP, STS-B, WNLI, QNLI, MPRC, RTS, SST-2, CoLA\}} validate the effectiveness of ISP wrt. several SOTA pruning methods. 
\end{itemize}

\section{Methodology}
\subsection{Revisiting LTH and Pre-trained Vision and Language Model Compression}
In the past few years, scaling neural networks to improve information absorption has been pivotal for good optimization and generalization performance, but this unbounded parameter growth has made them computationally expensive with excessive memory requirements. The trend undoubtedly continues with the recent forefront of transformers, where more and more layers are stacked with dense attention blocks (eg. T5 has $\sim10+$ billion parameters) calling for expensive computational resources and prolonged training or fine-tuning time. Recently, some work \cite{chen2021lottery,gan2022playing,chen2021chasing,chen2020earlybert,you2019drawing,prasanna2020bert} have explored Lottery Ticket Hypothesis (LTH) to understand the parameter redundancy in the current prevailing large scale transformer models, and attempted to compress it to non-trivial sparsities by repetitive initialization-training-pruning operation. Despite their success to find high-quality compressed models on a range of downstream tasks, it is impossible to ignore the cost of finding these subnetworks, since winning tickets can only be identified by pruning unimportant connections after fully training a dense network in a conventional LTH paradigm, which worsens significantly with increasing model size. For example, \cite{prasanna2020bert,chen2020lottery} explored pruning BERT to matching subnetworks at 40\% to 90\% sparsity across multiple tasks, on the other hand, \cite{gan2022playing} explored LTH for compressing large pre-trained VL models while preserving its performance.

\begin{figure}
  \centering
  \includegraphics[width=\linewidth]{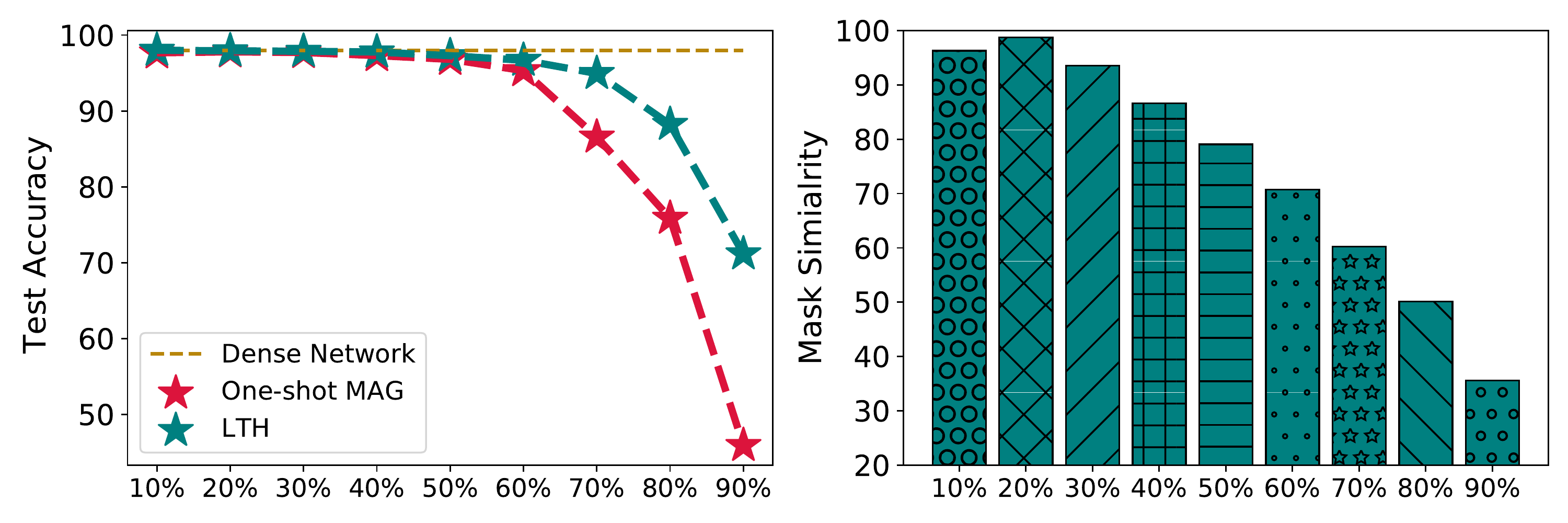}
  \vspace{-2em}
  \caption{Performance comparison of LTH and One-shot magnitude-based pruning of CLIP (\texttt{ViT-B32}) on \texttt{CIFAR-10} at sparsity $S \in \{10\%, 20\%, ..., 90\%\}$ (left). Cosine similarity between the binary prune masks obtained by LTH and One-shot magnitude pruning of CLIP (\texttt{ViT-B32}) on \texttt{CIFAR-10} (right).}
  \label{fig:mask_sim}
  \vspace{-2em}
\end{figure}

Recently, \cite{chen2020earlybert} explored the EarlyBird \cite{you2019drawing} idea and proposed jointly training BERT and some sparsity-inducing coefficients which can be used to draw the subnetworks followed by fine-tuning, but its joint training step again is as expensive as normal BERT training, and experimentally we found that its performance becomes sub-standard compared to LTH in non-trivial sparsity range. In this work, we propose a novel pruning strategy based on strong insights of the inherent benefits of gigantic size and model soups, which can be equivalent to or even better than LTH and its variant LTH-Rewind. We, for the first time, explore model compression for a recent extremely popular open vocabulary network CLIP along with BERT, to illustrate our approach benefits by using merely the computational cost equal to a single pass of conventional LTH.

\subsection{Instant Soup Pruning: A novel cost-effective pruning perspective}
In this section, we introduce a novel pruning algorithm, named \textbf{Instant Soup Pruning (ISP)} which primarily aims to reduce the computational overhead of conventional LTH while searching for lottery tickets in large-scale pre-trained transformers, facilitating benefits from both performance and computation perspective. 

ISP is motivated from the following three observations:
\vspace{-0.3cm}
\begin{itemize}
    \item \textit{Firstly}, large-scale pre-trained transformers are highly over-parameterized, and pruning them at trivial (eg. 10\%, 20\%, etc. depending on task and model size) sparsities does not require sophisticated pruning methods like LTH or LTH-Rewind to get high-quality sparse subnetworks. We surprisingly observed that \textit{at trivial sparsities, the sparse mask generated by LTH and cheap one-shot magnitude pruning is significantly similar} which thereby reflects in the test performance of the subnetworks. For example, Figure \ref{fig:mask_sim}(a) illustrates the performance of subnetworks obtained by LTH and one-shot pruning on CLIP (\texttt{ViT-B32}) and Figure \ref{fig:mask_sim}(b) illustrates the cosine similarity between the binary prune mask identified by LTH and one-shot pruning. It can be clearly observed that \textit{at trivial sparsities such as 10\%, 20\%, and 30\%, both unreasonably cheap one-shot pruning and expensive LTH identify approximately similar masks with 96.27\%, 98.75\%, and 94.32\% cosine similarity score}. It conveys a strong message to save the computation budget of full training passes of LTH, which are seemingly unnecessary.  

    \item \textit{Secondly}, Early-Bird \cite{you2019drawing} tickets, although limited to small architectures (ResNets, Vgg16, etc), conveyed a strong yet highly overlooked message that high-quality tickets can emerge at a very early training stage by pruning networks trained at much earlier points (before the accuracies reach their final top values). Recently, \cite{chen2020earlybert} showed that this observation holds true for BERT, but it cannot recover the full performance of LTH. To this end, complementary to our first observation, our work extends the early-bird findings by \textit{proposing to look more carefully while searching tickets at non-trivial sparsity (high sparsity regime) compared to trivial sparsity by progressively increasing training steps over each call to pruning routine} in the mask finding stage. 

    \item \textit{Lastly}, to mitigate the issue of pre-mature pruning to generate sub-standard pruning masks, we borrow inspiration from the intriguing phenomenon of “model soups”, which illustrate weights of large-scale independently fine-tuned models can be merged together into a better solution. Our work proposes a novel approach of \textit{mask soups by superposing multiple multiple cheap pruning masks, to attenuate the noise within them due to the presumably unstable state of the network while pruning}, giving a high-quality denoised pruning mask.  
\end{itemize}

\vspace{-0.5cm}
\paragraph{Algorithm Overview} Consider a dense, pre-trained network $f(x;\theta)$, as shown in Figure \ref{fig:main_architecture}, LTH trains $f$ to achieve minimum validation loss $f_{loss}$ using $E$ epochs with a test accuracy $f_{acc}$, when optimized with Adam optimizer on a training dataset $D$. Once, $f_{acc}$ is achieved, the fine-tuned network $f(x;\theta_{E})$ is pruned using magnitude pruning to generate a subnetwork $f(x; m \odot \theta_{E})$, with a mask $m \in \{0,1\}$. This process is repeated for $k$ iterations till the desired sparsity $S\%$ is achieved, generating a subnetwork $f_{LTH}(x;m \odot \theta_{\mathcal{O}(k \cdot E)})$ with accuracy $f^{LTH}_{acc}$. 

In contrast, our proposed approach ISP aims to generate $f_{ISP}(x;m'\odot\theta_{\mathcal{O}(E)})$ with accuracy $f^{ISP}_{acc}$, such that $f^{ISP}_{acc} \geq f^{LTH}_{acc}$ with sparse mask $m' \in \{0,1\}$ and sparsity $S\%$. Given the computational budget of $E$ epochs, which translates to $T$ steps with batch size $B$, ISP is composed of two distinct phases: \textit{mask generation phase} and \textit{fine-tuning phase} which uses $M$ and $(T-M)$ steps respectively to produce a high-quality fine-tuned subnetwork with desired sparsity $S\%$, usually outperforming LTH.

\label{alg:isp}
\begin{algorithm}
    \SetKwFunction{sparsity}{sparsity}
    \SetKwFunction{DenosiedPrune}{DenosiedPrune}
    \SetKwInOut{KwIn}{Input}
    \SetKwInOut{KwOut}{Output}

    \KwIn{Pre-trained Network: $f(x;\theta_{0})$; Initial Training Seed: $t$; Compression ratio: $s$\%; Desired sparsity: $S\%$; Training Budget: $T$}
    \KwOut{Pruned trained subnetwork with a mask $m' \in \{0,1\}$: $f_{ISP}(x;m'\odot\theta_{\mathcal{O}(E)})$ }
    
    \tcc{$k$ is chosen st. $\sum_{i = 0}^{k}(i + 1) \cdot t \leq M$}
    \For{$i \leftarrow 0$ \KwTo $k$}{ 

        \tcc{Sparsity dependent training steps before call to pruning routine. Spend low training steps for trivial sparsity compared to non-trivial sparsity.}
        Train $f(x;\theta_{i \cdot t}) \rightarrow f(x;\theta_{(i + 1) \cdot t})$
        
        \If{$\sparsity(f) < S\%$}{
            $m_{(i+1)} \leftarrow \DenosiedPrune(f, s, m_i)$
            \tcc{Compress network with newly returned denoised mask}
            Apply $m_{(i+1)} \rightarrow f(x;\theta_{(i + 1) \cdot t})$
        }
    }

    \tcc{Fine-tune the obtained subnetwork for the remaining $(T - ((k + 1) \cdot t))$ steps in  computational budget.}
    Fine-tune $f(x;\theta_{(k + 1) \cdot t}) \rightarrow f(x;\theta_{T - ((k + 1) \cdot t)})$
    \caption{Instant Soup Pruning}
\end{algorithm}

\subsubsection{Mask Generation Stage} We will first discuss our novel, computationally efficient, and high-quality sparse mask generation steps for ISP incorporating the aforementioned motivation. Note that similar to IMP, ISP is also an iterative train-prune-retrain procedure, but it uses a highly optimized train/re-train subroutine starting from the parameter state obtained from the previous iteration. Given the computational budget of $M$ steps, the mask generation stage of ISP is carefully designed to start in a relaxed fashion (spending few learning steps) while pruning at trivial sparsities and gradually become meticulous while approaching non-trivial sparsity regime, acknowledging the sensitivity of pruning while operating in high sparsities.  

More specifically, we introduce a hyperparameter $t$, which is defined as a small initial seed step count usually equates to the number of steps required to look $\sim10\%$ of training data with batch size $B$, before the first call to pruning routine. At $i$-th call to the pruning routine, we use $t \times (i + 1)$ training steps for calibrating (training) the network before pruning. For example, at the $0$-th iteration when the network is dense, ISP uses steps equivalent to 10\% of training data while at $5$-th iteration, it uses 60\% of training data in the re-training. 

To ensure that ISP does not produce sub-standard quality mask at each pruning iteration, we next provide details of our novel \textbf{denoised pruning routine} which is inspired by recently proposed ``model soups" phenomenon, to average out noise induced due to pre-mature pruning.

\vspace{-0.1cm}
\begin{algorithm}
    \SetKwFunction{MagnitudePruning}{MagnitudePruning}
    \SetKwFunction{DenosiedPrune}{DenosiedPrune}
    
    \SetKwInOut{KwIn}{Input}
    \SetKwInOut{KwOut}{Output}

    \KwIn{Network to prune: $f(x; m_i \odot \theta)$ where $m_i$ denotes binary mask at $i$-th iteration of ISP; Denoiser Count: $N$; Compression Rate: $s$\%; Look-ahead steps: $C$ (typically $10 \leq C \leq 100$ steps) }
    \KwOut{New denoised mask:  $m_{(i + 1)}$}
    $m_{temp} \leftarrow \MagnitudePruning(f(x;\theta), s\%)$
  
    \For{$n \leftarrow 0$ \KwTo $N$}{ 

        \tcc{Look-ahead training to generate multiple masks for denoising}
        $P_{n} \leftarrow $ Training protocol \tcp{learning rate, weight decay, etc.}
        $D_{n} \leftarrow $ Random data samples
        
        Look-ahead training: $f(x;\theta) \rightarrow f(x;\theta_{C})$

        \tcc{Denosing to average mask noise}

        $m_{temp} \leftarrow m_{temp} \cup \MagnitudePruning(f(x;\theta_{C}), s\%)$
    }

    $m_{(i + 1)} \leftarrow \texttt{One-shot-adjustment}(m_{temp}, f(x;\theta))$
    
    \KwRet{$m_{(i + 1)}$}
    \label{alg:denoising}
    \caption{Denoised Pruning Procedure}
\end{algorithm}

\vspace{-0.4cm}
\paragraph{Denoised Pruning:} Recently, several works \cite{wortsman2022model,ilharco2022patching,juneja2022linear} have validated the intriguing phenomenon of model soups, which suggest training multiple models with various hyperparameters and average the weights of models fine-tuned independently at no cost to achieve comparatively high performance. Motivated by their argument that fine-tuned models optimized independently from the same pre-trained initialization lies in the same basin of the error landscape and averaging them improves generalization, we are tempted to ask an unexplored question: \textit{Can we generate extremely cheap pruning masks using varying hyperparameters and average them out to improve the quality?}

To this end, we propose a novel \textbf{Denoised Pruning Procedure}, which explores the superimposition of computationally cheap pruning mask obtained by magnitude-based one-shot pruning of the network with marginal look-ahead training. Algorithm 2 illustrates the details of our denoising procedure which perform look-ahead training of the network using varying training protocol for a random subset of training samples, merely using $10-100$ training steps, to generate $N$ candidate binary masks. In order to superimpose them, we found that a simple \underline{union} of these $N$ binary masks is sufficient to improve the quality, facilitating ISP to achieve comparable/even better performance than expensive LTH.  We hypothesize (and later experimentally validate) that our denoising procedure significantly helps eliminate any induced noise due to pre-mature pruning during ISP iterations, specifically at non-trivial sparsities.

\subsubsection{Fine-tuning Stage} Our proposed method (ISP), is a single pass pruning algorithm that generates a high-quality fine-tuned subnetwork with desired sparsity $S\%$ using the computational budget equivalent to a single pass of LTH ($T$ steps). As mentioned before, ISP is composed of the mask generation phase ($M$ steps) followed by fine-tuning the obtained subnetwork for the remaining ($T-M$ steps). Note that, unlike conventional LTH, ISP does not restart the network training from the initial pre-trained weight; instead, as explained in Algorithm 1, it simply fine-tunes the network state (unchanged optimizer, learning rate, etc.) obtained immediately after pruning $S\%$ of parameters for $(T-M)$ steps.

\subsection{Instant Model Soup: A sparsity-inspired extension to dense training}
In the conventional setting, to improve the generalization performance of a model, it is highly recommended to train multiple models and use their ensemble on the test set, but it comes at a heavy inference and training cost. Recently, several works \cite{wortsman2022model,ilharco2022patching,juneja2022linear} have illustrated that, unlike conventional ensembles, the weights of multiple fine-tuned large pre-trained models can be merged together by simply averaging their weights (aka. model soup) to beat the performance of ensembles without incurring any additional inference or memory cost. While this approach effectively reduces the additional inference overhead of ensembles, it still requires expensive fine-tuning of multiple large pre-trained models with varying hyperparameters. Inspired by our idea of Instant Soup Pruning, which illustrates that binary masks originated from cheap training with different training protocols can be superimposed/denoised to improve quality, we are enticed to explore: \textit{Can we denoise the initial pre-trained weights of dense large transformers using sparse cheap training to inject the model soup benefits early during fine-tuning, eliminating the need to generate multiple fully fine-tuned models for model soups?} 

\label{alg:denoising}
\begin{algorithm}
    \SetKwFunction{MagnitudePruning}{MagnitudePruning}
    \SetKwFunction{Interpolate}{Interpolate}
    \SetKwFunction{DeepCopy}{DeepCopy}
    \SetKwFunction{Train}{Train}
    \SetKwFunction{sample}{sample}
 
    
    \SetKwInOut{KwIn}{Input}
    \SetKwInOut{KwOut}{Output}

    \KwIn{Pre-trained Model:  $f(x; \theta_0)$; $D_{Train}$; $D_{Val}$; Denosier Count: $K$}
    \KwOut{New denoised Pre-trained Model:  $f(x; \theta_{new})$}
    $f(x; \theta_{interpolated}) \leftarrow \DeepCopy(f(x; \theta_0))$
    
    \For{$k \leftarrow 0$ \KwTo $K$}{ 

        \tcc{Prune to create a sparse subnetwork from input model}
        $f(x; m_s \odot \theta_0) \leftarrow \MagnitudePruning(f(x; \theta_0), s\%)$
        
        \tcc{Weakly train sparse subnetwork for denoising}
        $H_k, C_k \leftarrow $ Training Protocol, Steps 
        
        $D_k \leftarrow \sample(D_{Train})$
        
        $f_{weak}(x; m_s \odot \theta_k) \leftarrow \Train(f(x; m_s \odot \theta_0), H_k, C_k, D_k)$

        \tcc{Denoising using linear interpolation with val data}

        $f(x; \theta_{interpolated}) \leftarrow \Interpolate(f_{weak}(x; m_s \odot \theta_k), f(x; \theta_{interpolated}), D_{Val})$
    }

    $f(x; \theta_{new}) \leftarrow f(x; \theta_{interpolated})$
    
    \KwRet{$f(x; \theta_{new})$}
    
    \caption{Instant Model Soup}
    \label{alg:model_soup}
\end{algorithm}
\vspace{-0.2cm}
Our primary objective is to ripe the benefits of model soups without the need to generate multiple fully fine-tuned models, thereby contenting the hurdle of high training overhead of model soups. Algorithm \ref{alg:model_soup} provide details of the pseudocode of our new \textbf{Instant Model Soup (IMS)} approach that uses cheap sparse training for injecting model soup benefits at marginal cost early before fine-tuning to save the hurdle of multiple finetuning and averaging. More specifically, IMS first creates multiple subnetworks with varying sparsities from the pre-trained dense model and train them independently for a few iterations ($\sim$100 iterations) using different hyperparameter configuration, and data subsets. Next, all the weakly trained subnetwork weights are merged together with the initial pre-trained model weights using linear interpolation following \cite{ilharco2022patching}. Our proposed usage of sparse subnetworks for denoising significantly reduces the computational cost of the dense training step, thereby making it more efficient. Our experiments on CLIP (\texttt{ViT-B32}) and BERT-\texttt{BASE} impart an interesting finding that pre-trained models denoised by IMS can achieve performance comparable to model soups, without expensive full fine-tuning and then averaging.


\vspace{-0.1cm}
\begin{table}[h]
\centering
\resizebox{0.47\textwidth}{!}
{\begin{tabular}{lcccccc}
\toprule
& Initial LR & Epochs & Compression Rate & Look Ahead ($C$) & Weight Decay & Denoiser \\
\midrule
BERT$_\texttt{BASE}$ & 2 $\times$ 10$^{-5}$ & 10 & 10\% & 30 iter & 0.0-AdamW & 4\\
CLIP$_\texttt{ViT-B32}$ & 1 $\times$ 10$^{-5}$ & 22 & 15\% & 50 iter & 0.1-AdamW & 5\\
\bottomrule
\end{tabular}}
\caption{Details of our primary hyperparameter configurations used in our experiments across different evaluation datasets.} 
\vspace{-0.4cm}
\label{tab:hyperparameter}
\end{table}

\begin{table*}[h]
\centering
\caption{Details of fine-tuning CLIP (\texttt{ViT-B32}) at varying sparsity levels using Instant Soup Pruning following the settings listed in Table \ref{tab:hyperparameter}. Learning rate decays linearly from the initial value to zero. The evaluation metrics follow standards in \cite{radford2021learning}. Entries with errors are the average across three runs, and errors are the standard deviations. LTH results are obtained using IMP.}
\resizebox{1\textwidth}{!}{\begin{tabular}{l|ccc|ccc|ccc|ccc|ccc|ccc}
\toprule
\multirow{2}{*}{Pruning Method} & \multicolumn{3}{c|}{Cars} & \multicolumn{3}{c|}{MNIST} & \multicolumn{3}{c|}{SVHN} & \multicolumn{3}{c|}{GTSRB}& \multicolumn{3}{c|}{CIFAR10} & \multicolumn{3}{c}{CIFAR100}\\ \cmidrule{2-19}
& 30\% & 40\% & 50\% & 70\% & 80\% & 90\% & 70\% & 80\% & 90\% & 50\% & 60\% & 70\%&  60\% & 70\% & 80\% & 60\% & 70\% & 80\%\\
\midrule
\rowcolor[gray]{0.9} 
\multirow{1}{*}{Full CLIP$_\texttt{ViT-B32}$} & \multicolumn{3}{c|}{76.43 $\pm$ 0.5} & \multicolumn{3}{c|}{99.61 $\pm$ 0.07} & \multicolumn{3}{c|}{97.40 $\pm$ 0.11} & \multicolumn{3}{c|}{99.08 $\pm$ 0.24}& \multicolumn{3}{c|}{97.6 $\pm$ 0.24}& \multicolumn{3}{c}{89.35 $\pm$ 0.19}\\
\midrule
Random & 9.11 & 5.58 & 4.47 & 98.71 & 97.42 & 87.04 & 89.85 & 85.61 & 73.74 & 93.76 & 93.65 & 90.97 & 74.51 & 69.84 & 64.96 & 45.20 & 39.92 & 43.24\\
One-shot [Mag] &  71.95 & 68.07 & 56.79 &  99.27 & 98.87 & 97.47 & 95.02 & 91.96 & 85.76 &  98.60 & 97.84 & 96.14 & 95.31 & 86.56 & 75.85 & 80.39 & 63.18 & 46.63  \\
Progressive [Mag] & 69.85 & 68.62 & 64.43 & 99.52& 97.77& 95.19& 95.78 & 90.53 & 85.75 & 98.97 &  97.57& 96.34 & 95.25 & 90.87 & 78.11 & 81.79 & 70.53 & 60.93\\
EarlyBird \cite{you2019drawing} & 72.53 & 70.76 & 65.90 & 99.38 & 98.96 & 97.64 & 96.34 & 95.93 & 87.02 & 98.15 & 98.19 & 97.26 & 96.06 & 94.18 & 86.84 & 84.22 & 76.79 & 65.67\\
\midrule
SNIP \cite{lee2018snip} & 71.51 & 68.79 & 59.01 & 99.25 & 98.72 & 97.50  & 95.33 & 91.94 & 82.98 & 98.62 & 97.95 & 96.22& 95.01 & 87.45 & 76.12 & 81.10 & 62.89 & 55.89\\
GraSP \cite{wang2020gradient} & 71.42 & 68.55 & 58.12 & 99.30 & 98.51 & 97.15 & 95.09 & 91.44 & 84.72 & 98.37 & 97.42 & 95.91 & 95.20 & 86.89 & 75.88 &80.67 & 66.31 & 52.30\\
\midrule
LTH \cite{frankle2018lottery} & 73.97 & 72.02 & 66.12 & 99.41 & 99.38 & 98.22 & 96.69 & 95.28 & 87.41 & 98.71 & 98.35 & 97.79 & 96.42 & 94.91 & 87.47 & 84.25 & 78.60 & 65.38\\
LTH - Rewind & 74.28 & 72.09 & 66.07 & 99.62 & \textbf{99.64} & 98.18 & 96.72 & 95.22 & 87.47 & 98.78 & 98.36 & 97.87 & 96.53 & 94.88 & 87.28 & 84.46 & 78.62 & 65.71\\

Lottery Pool \cite{yin2022lottery} & 73.10 & 70.53 & 64.67 & 99.25 & 98.97 & 97.76 & 96.54 & 95.12 & 87.29 & 98.52 & 98.30 & 97.55 & 96.14 & 94.50 & 87.11 & 84.07 & 78.21 & 64.39 \\
\midrule
\rowcolor[gray]{0.9} 
ISP [Ours] & \textbf{75.13} & \textbf{72.20} & \textbf{66.32}& \textbf{99.69} & 99.61 & \textbf{98.82} & \textbf{96.93} & \textbf{96.46} & \textbf{87.59} & \textbf{99.06} & \textbf{99.01} & \textbf{98.52} & \textbf{96.82} & \textbf{95.18} & \textbf{91.20} & \textbf{85.11} & \textbf{79.57} & \textbf{71.09} \\
\rowcolor[gray]{0.9}
\ \ \ \ \ \ (std.) & $\pm$0.34 & $\pm$0.27 & $\pm$0.82 & $\pm$0.07 & $\pm$0.15 & $\pm$0.21 & $\pm$0.08 & $\pm$0.05 & $\pm$0.11 & $\pm$0.15 & $\pm$0.29 & $\pm$0.32 & $\pm$0.15 & $\pm$0.20 & $\pm$0.14 & $\pm$0.19 & $\pm$0.22 & $\pm$0.17\\

\bottomrule
\end{tabular}}
\vspace{-0.6em}
\label{tab:pretrain_clip}
\end{table*}


\section{Experiments and Analysis}
\subsection{Network, Dataset, and Settings} In our experiments, we adopted the official CLIP implementation provided by \cite{radford2021learning} as our starting point for our experiments using pre-trained vision transformer \cite{dosovitskiy2020image} (\texttt{ViT-B32}) models.  For fine-tuning CLIP, we use the frozen final classification layer output by CLIP's text tower to eliminate the necessity of introducing any learnable parameters while in the case of BERT, we add a final task-specific classification layer($\sim 3\%$ of total parameter count). For our BERT-related experiments, we use the HuggingFace \cite{wolf2019huggingface} pre-trained weights of BERT$_{\texttt{BASE}}$ transformer blocks and hidden state size 768. Additional necessary details of our hyperparameters required for fine-tuning are provided in Table \ref{tab:hyperparameter}. Note that during pruning, we prune the key trainable part of the network \textit{(key, query, value, dense)} ignoring embeddings for simplicity. We consider a diverse set of image classification tasks from \cite{radford2021learning}: \texttt{Cars, GTSRB, MNIST, SVHN, CIFAR10/100} and downstream NLP tasks from GLUE \cite{wang2018glue} benchmark: \texttt{MNLI, QQP, STS-B, WNLI, QNLI, MPRC, RTS, SST-2, CoLA}; to thoroughly investigate the effectiveness of our proposed approaches wrt. state-of-the-art pruning benchmarks.  In addition to Lottery tickets, we have compared Instant Soup Pruning (ISP) against several recently proposed pruning methods such as Lottery Pools \cite{yin2022lottery}, Early bird \cite{you2019drawing}, two popular pruning at initialization methods (SNIP \cite{lee2018snip}, GraSP \cite{yu2020gradient}), Progressive pruning (Iterative pruning and training), and one-shot magnitude pruning.

\begin{table*}[h]
\centering
\caption{Details of fine-tuning BERT (\texttt{BASE}) at varying sparsity levels using Instant Soup Pruning following the settings listed in Table \ref{tab:hyperparameter}. Learning rate decays linearly from the initial value to zero. The evaluation metrics follow standards in \cite{wolf2019huggingface}. Entries with errors are the average across three runs, and errors are the standard deviations. LTH results are obtained using IMP.} 
\resizebox{0.95\textwidth}{!}{\begin{tabular}{l|ccccccccc}
\toprule
Dataset & MNLI & QQP & STS-B & WNLI & QNLI & MPRC & RTS & SST-2 & CoLA \\
\midrule
Sparsity & 70\% & 90\% & 50\% & 90\% & 70\% & 50\% & 60\% & 60\% & 50\% \\
\midrule
\rowcolor[gray]{0.9} 
Full BERT$_\texttt{BASE}$ & 82.4 $\pm$ 0.5 & 90.2 $\pm$ 0.5 & 88.4 $\pm$ 0.3 & 54.9 $\pm$ 1.2 & 89.1 $\pm$ 1.0 & 85.2 $\pm$ 0.1 & 66.2 $\pm$ 3.6 & 92.1 $\pm$ 0.1 & 54.5 $\pm$ 0.4 \\
Random & 67.5  & 76.3 & 21.0 & 53.5 & 61.9 & 69.6 & 56.0 & 83.1 & 9.6 \\
One-shot & 78.8 & 86.2 & 83.9 & 53.1 & 86.2 & 83.7 & 62.9 & 86.5 & 49.7\\
Progressive  & 79.1 & 87.5 & 85.0 & 53.3 & 87.2 & 83.8 &  65.4 & 86.6 & 52.2\\
EarlyBird  & 82.5 & 89.4 & 88.1 & 54.0 & 88.5 & 84.6 & 66.1 & 91.2 & 53.5\\
\midrule
Lottery Ticket & 82.6 & 90.0 & 88.2 & 54.9 & 88.9 & 84.9 & 65.0 & 91.9 & 53.8 \\
Lottery Pool & 80.4 & 89.1 & 86.4 & 50.9 & 87.6 & 84.5 & 62.7 & 90.9 & 52.6\\
\midrule
\rowcolor[gray]{0.9} 
ISP & \textbf{82.71 $\pm$ 0.6} & \textbf{90.59 $\pm$ 0.5} & \textbf{88.64 $\pm$ 0.1} & \textbf{55.33 $\pm$ 0.3} & \textbf{90.06 $\pm$1.0} & \textbf{85.38 $\pm$ 0.1} & \textbf{65.96 $\pm$ 0.3} & \textbf{92.43 $\pm$ 0.6} & 53.61 $\pm$ 0.2\\
\bottomrule
\end{tabular}}
\vspace{-0.5em}
\label{tab:pretrain_bert}
\end{table*}

\subsection{Performance comparison of Instant Soup Pruning wrt. SOTA pruning methods}
In this section, we conduct a systematic and extensive study to understand the performance benefits of our proposed Instant Soup Pruning  in terms of fine-tuning accuracy vs. pruning ratios by comparing against multiple state-of-the-art pruning methods. We first test our approach on recently proposed CLIP (ViT-B32) \cite{radford2021learning} (unexplored for pruning till date) pre-trained with contrastive supervision from image-text pairs. For effective comparison and simplicity, all baselines and our proposed approach ISP are trained with similar optimizer and training settings provided in Table \ref{tab:hyperparameter}. Our EarlyBird \cite{you2019drawing}, SNIP \cite{lee2018snip}, and GraSP \cite{yu2020gradient} baselines closely follow the implementation provided by their official GitHub repositories. In addition, to distinguish ISP from conventional progressive pruning, our progressive pruning baseline is implemented as periodic pruning using magnitude-based pruning followed by retraining. Lottery pools \cite{yin2022lottery} is an interesting way to merge LTH by-product tickets. To ensure that the sparsity ratio remains comparable to ISP, we further prune the merged tickets obtained by pooling to the required sparsity, for reporting performance.

Our results for CLIP are summarized in Table \ref{tab:pretrain_clip}. We first observe that among all baselines for CLIP compression, LTH consistently performs better, and rewinding helps in further improving its performance. We found that among recent pruning at initialization methods (SNIP and GraSP), SNIP has comparatively better performance than GraSP and they tend to be slightly better than one-shot magnitude pruning. It can be clearly observed that ISP can beat expensive LTH (including rewinding) as well as all other baselines for almost all benchmark datasets and sparsity ratios. Very interestingly, for CIFAR-10 and CIFAR-100, we found that performance improvement of ISP increases with the sparsity ratio. For example, ISP surprisingly outperforms LTH-rewind by \underline{$\sim3.92$ and $\sim5.38\%$ on CIFAR-10 and CIFAR-100} respectively, while consuming merely fine-tuning cost equivalent to one single pass of LTH without the necessity of bookkeeping the rewinding weights of LTH-rewind. Our experimental results for BERT-base are summarized in Table \ref{tab:pretrain_bert}. For BERT-related experiments, we have replicated the setting in \cite{chen2020lottery} and reported the performance of ISP across various GLUE benchmarking datasets at the sparsity level where LTH is able to identify winning tickets. Note that ISP is able to comfortably \underline{outperform LTH across 8 out of 9 tasks} (noticeably for QNLI where ISP beat LTH by $>1\%$).

\subsection{Analysis of Denoising Iterations in ISP}
Our proposed approach ISP is augmented by a novel idea of mask denoising, which explores the superimposition of computationally cheap pruning mask obtained by magnitude-based one-shot pruning of the network with marginal look-ahead training. In this section, we try to investigate the implication of our denoising iterations in improving ISP performance. Table \ref{tab:denoiser} summarizes the performance comparison of ISP with/without the mask denoising while keeping the training settings exactly the same. Across both candidate architectures (CLIP and BERT), it can be clearly observed that ISP performance is significantly boosted by replacing the simple one-shot pruning with our denoise pruning routine. In addition, we also investigated how the number of denoising iterations will impact the ISP performance (see Table \ref{tab:denoiser_count}) and found that 4-5 denoising steps are sufficient for the denoised pruning, and increasing them beyond that does not provide a very noticeable performance gain. For consistency, in CLIP-related experiments, we have used 5 denoising iterations while for BERT, our results are reported using 4 denoising iterations.

\begin{table}
\centering
\caption{Impact of Denoising Module in improving the performance of ISP. Results are reported for three independent runs.} 
\resizebox{0.45\textwidth}{!}{\begin{tabular}{lcccc}
\toprule
\multirow{2}{*}{Approach} & \multicolumn{2}{c|}{CLIP$_\texttt{ViT-B32}$} & \multicolumn{2}{c}{BERT$_\texttt{BASE}$} \\ \cmidrule{2-5}
& SVHN & CIFAR-100 & QQP & QNLI \\
\midrule
ISP - Denoiser &  96.11$\pm$0.21 & 70.32$\pm$0.13 & 89.96$\pm$0.39 & 89.28$\pm$0.87\\
\rowcolor[gray]{0.9}
\midrule
ISP (Ours) & 96.46$\pm$0.05 & 71.09$\pm$0.17& 90.59$\pm$0.47 & 90.06$\pm$1.01\\
\bottomrule
\end{tabular}}
\vspace{-1.4em}
\label{tab:denoiser}
\end{table}

\begin{table}
\centering
\caption{Performance comparison of ISP wrt. denoiser count on CIFAR10 with CLIP (ViT-B32) pruned at 80\% sparsity.} 
\resizebox{0.44\textwidth}{!}{\begin{tabular}{c|cccccc}
\toprule
Denoiser Count & 0 & 2 & 4 & 6 & 8 & 16\\
\midrule
\rowcolor[gray]{0.9}
Performance & 70.32 & 70.95 & 71.15 & 71.16 & 71.07 & 71.16\\
\bottomrule
\end{tabular}}
\vspace{-1.4em}
\label{tab:denoiser_count}
\end{table}

\subsection{Understanding the benefits of Instant Model Soups for pre-trained models}
In this section, we discuss the benefits of our sparsity-inspired extension, Instant Model Soup (IMS), and experimentally validate its surprising ability to improve the quality of  pre-trained models at marginal cost. Unlike model soups, IMS provides a unique opportunity to eliminate the requirement to generate multiple fully fine-tuned models to average, thereby restricting the computational complexity equivalent to the cost of fine-tuning a single model. Table \ref{tab:ims} illustrates the performance comparison of IMS with respect to two model soup variants (uniform and greedy) proposed in \cite{wortsman2022model}. Note that uniform and greedy soups results are generated using the amalgamation of 8 independent models fine-tuned till the final accuracy with different hyperparameters. Our experiments across CLIP and BERT illustrate that  by carefully fine-tuning IMS, it is surprisingly possible to comfortably beat the model soup variants significantly. The denoised pre-trained model generated by IMS has the ability to converge to equivalent (even better) performance than model soups. Adhering to the theme of ISP, IMS also conveys a strong message that it is not necessarily important to wait till model convergence to ripe the benefits of soup, but astonishingly soup benefits are available to ripe early during the fine-tuning at a marginal computational cost. 

\begin{table}
\centering 
\caption{Fine-tuning performance comparison of our proposed approach (IMS) wrt. basic fine-tuning and model soup variants.}
\resizebox{0.47\textwidth}{!}{\begin{tabular}{lccccc}
\toprule
\multirow{2}{*}{Approach} & \multicolumn{3}{c|}{CLIP$_\texttt{ViT-B32}$} & \multicolumn{2}{c}{BERT$_\texttt{BASE}$} \\ \cmidrule{2-6}
& Cars & CIFAR10 & CIFAR100 & MNLI & QNLI \\
\midrule
Pretrained-\texttt{BASE} &  76.43 & 97.60  & 89.35 & 82.39 & 90.04\\
Uniform Soup \cite{wortsman2022model} &  76.32 & 97.68 & 89.20 & 82.41 & 89.76\\
Greedy Soup \cite{wortsman2022model} &  77.95 & 98.05 & 89.54 & 83.01 & 90.64\\
\rowcolor[gray]{0.9}
\midrule
IMS [Ours]& 78.79 & 98.01  & 89.64 & 83.63 & 91.23\\
\rowcolor[gray]{0.9}
\ \ \ (std.) & $\pm$ 0.32 & $\pm$ 0.07 & $\pm$ 0.12& $\pm$ 0.43 & $\pm$ 0.19\\
\bottomrule
\end{tabular}}

\vspace{-0.4cm}
\label{tab:ims}
\end{table}

\begin{table}
\centering
\caption{Performance comparison of ISP wrt. look-ahead on CIFAR100 with CLIP (ViT-B32) pruned at 60\% sparsity.} 
\resizebox{0.44\textwidth}{!}{\begin{tabular}{c|cccccc}
\toprule
Look-ahead Count(C) & 10 & 30 & 50 & 100 & 150 & 200\\
\midrule
\rowcolor[gray]{0.9}
Performance & 84.71 & 84.92 & 85.18 & 85.19 & 84.41 & 85.41\\
\bottomrule
\end{tabular}}
\vspace{-0.4cm}
\label{tab:look_ahead}
\end{table}

\section{Related Work}
Linear interpolation of neural network weights has recently achieved significant attention, but due to numerous nonlinear activations within a neural network, it is still debatable if linearly interpolating between two sets of weights can result in a high accuracy solution. Recently, \cite{frankle2020linear,nagarajan2019uniform,von2020neural,matena2021merging,wortsman2022model,wortsman2022robust,choshen2022fusing,izmailov2018averaging,neyshabur2020being} have studied the interpolation of deep networks and validated performance benefits when training starts from a common initialization or some segment of the optimization trajectories are shared. While \cite{nagarajan2019uniform,frankle2020linear} focused on mergability in the case of models trained on a single task, \cite{wortsman2022robust} found that weight interpolation can not only benefit fine-tuning tasks but also under distribution shift. More specifically, they average zero-shot and fine-tuned models, finding improvements in- and out-of-distribution. 

Recently, \cite{matena2021merging} used Fisher-weighted averaging of language models before and after fine-tuning on downstream tasks. They merged models with the same pre-trained initialization that
are fine-tuned on different text classification tasks. In the late phases of training, \cite{von2020neural} studied making copies of a subset of the neural network parameters and proposed to independently optimize them, followed by averaging. Moreover, \cite{wortsman2022model} proposed to average fine-tuned models across independent runs with hyperparameter diversity, modifying all the weights of the network,  and showing significant performance benefits. In addition to model weight averaging, \cite{bansal2021revisiting, yang2022deep} explored the idea of model stitching, where given two trained and frozen models A and B, a ``stitched model" formed by connecting the bottom-layers of A to the top-layers of B, with a simple trainable layer between them. \cite{fort2019deep} studied deep ensembles which have empirically shown promise for improving the accuracy, uncertainty and out-of-distribution robustness of deep learning models.

\vspace{-0.3cm}
\section{Conclusion}

In this work, we introduced Instant Soup Pruning, a model soup-inspired perspective dedicated to generating LTH quality subnetworks, using a fraction of the original IMP cost. ISP is augmented by a denoising pruning module which helps in replacing the expensive intermediate pruning stages of IMP with computationally efficient weak mask generation and aggregation routine. Additionally, we present Instant Model Soup, which provides an opportunity to inject the benefits of model soups in dense pre-trained models at marginal training cost, thereby improving fine-tuning performance comparable to model soups. 
Our future work will aim for a more theoretical understanding of the role of
our denoisers in providing experimental benefits.

\section{Acknowledgement}
The research is based upon work supported in part by the Intelligence Advanced Research Projects Activity (IARPA) under Contract No. 2022-21102100004. We also acknowledge support from the National Science Foundation AI Center Institute for Foundations of Machine Learning (IFML) at the University of Texas at Austin.

\bibliography{example_paper}
\bibliographystyle{icml2023}



\end{document}